\documentclass[11pt,letterpaper]{article}
\usepackage{arxiv}
\usepackage[utf8]{inputenc}
\usepackage[T1]{fontenc}
\usepackage{lmodern}
\usepackage{mathptmx}
\usepackage{amsmath,amssymb}
\usepackage{graphicx}
\usepackage{tikz}
\usetikzlibrary{arrows.meta,positioning}
\usepackage{array,calc,longtable,booktabs}
\usepackage{microtype}
\usepackage{caption}
\usepackage[authoryear,round]{natbib}
\usepackage{xurl}
\usepackage[hidelinks,unicode]{hyperref}
\hypersetup{pdftitle={Man and Machine: Artificial Intelligence and Judicial Decision Making},pdfauthor={Arthur Dyevre and Ahmad Shahvaroughi}}
\title{Man and Machine: Artificial Intelligence and Judicial Decision Making}
\author{%
  \begin{tabular}[t]{c}
    \textbf{Arthur Dyevre}\thanks{Corresponding author.}\\
    Leuven Centre for Empirical Jurisprudence\\
    \texttt{artdyevre@gmail.com}
  \end{tabular}%
  \hspace{1.2em}%
  \begin{tabular}[t]{c}
    \textbf{Ahmad Shahvaroughi}\\
    Leuven Centre for Empirical Jurisprudence\\
    Leuven Laboratory for\\
    Experimental Social Psychology\\
    \texttt{ahmad.shahvaroughi@kuleuven.be}
  \end{tabular}%
}
\date{}

\begin{document}
\maketitle

\begin{abstract}
The integration of artificial intelligence (AI) into judicial decision
making -- particularly in pretrial, sentencing, and parole contexts --
has generated a substantial and rapidly growing literature. Across
computer science, economics, law, criminology, and psychology,
researchers have examined the reliability, fairness, and real-world
effects of AI-assisted decision making. Yet this literature remains
fragmented, and differences in assumptions, concepts, and research
priorities make it difficult to assess what is actually known. Using
criminal justice risk assessment as a focal case, this article makes two
contributions. First, we develop a conceptual framework that
distinguishes and relates three central questions: (1) the predictive
validity of automated risk assessment tools; (2) how algorithmic risk
assessments compare with human predictions (AI-versus-Human); and (3)
how algorithmic recommendations affect judges' decisions
(AI-plus-Human). Second, we use this framework to synthesize the
empirical evidence addressing each of these questions. Our review
identifies important limitations in existing research on predictive
validity, as well as substantial gaps in understanding how judges
respond to AI advice and how those responses vary across individuals and
decision-making environments. The available evidence suggests that AI
decision aids have, so far, had at most modest effects on pretrial and
sentencing decisions. We conclude that further research is needed to
understand how judges make decisions in noisy informational environments
and under what conditions AI tools can produce meaningful improvements
in judicial decision making.
\end{abstract}

\section{Introduction}

Artificial intelligence (AI) technologies are increasingly used or
proposed as decision aids in legal and judicial settings. In criminal
justice, predictive tools are employed to estimate risks such as failure
to appear, recidivism, or violation of parole conditions and to inform
decisions concerning pretrial release, sentencing, and parole. In the
United States, instruments such as COMPAS and the Public Safety
Assessment (PSA) have become prominent examples of this development.
Similar technologies have prompted extensive debate over whether
algorithmic systems can improve the consistency, accuracy, and fairness
of judicial decision making or, instead, reproduce and amplify
shortcomings already embedded in criminal justice institutions
\citep{Engel2025z7013,Niblett2024z7016,Fazel2022z7093,Gursoy2022z7009,Salo2019z7091,StephenWormith2018z7224}.

These debates have generated a large empirical literature spanning
computer science, economics, law, criminology, psychology, and related
fields. Yet the resulting knowledge remains fragmented. Different
disciplines tend to emphasize different questions and apply different
assumptions about data and the data generation process. Data scientists
and criminologists commonly evaluate the predictive performance and
fairness of risk assessment instruments \citep{Fazel2022z7093}. Economists and
computer scientists have compared algorithmic predictions with the
decisions of human judges and developed methods to address the difficult
counterfactual problems such comparisons entail \citep{Angelova2025z7253,Kleinberg2018z5972}.
Psychologists and behavioural researchers have examined the biases,
heuristics, and informational limitations of human decision makers, as
well as the ways in which people respond to algorithmic advice
\citep{GrgicHlaca2019z7155,Rachlinski2017z6997}. Although these literatures address closely connected
problems, differences in terminology, assumptions and priorities often
obscure what, collectively, they tell us about the actual effects of AI
in judicial decision making.

This article seeks to provide an integrated account. Using criminal
justice risk assessment as a focal case, we make two contributions.
First, we develop a conceptual framework that distinguishes three
questions: predictive validity, AI-versus-Human, and AI-plus-Human. The
first concerns the performance of the algorithmic tool itself: how well
does it predict the outcome it is designed to predict, and how robustly
does that performance generalize across settings and groups? The second
is comparative: how does algorithmic prediction perform relative to
unaided human judgment? The third concerns decision aid in practice:
what happens when an algorithmic recommendation is introduced into a
decision process that ultimately remains controlled by a human judge?
These questions are related, but answers to one do not determine answers
to the others. A risk assessment instrument may possess predictive
validity without outperforming judges; it may outperform judges when
considered in isolation without improving decisions when used as advice;
and even a highly accurate recommendation may have little practical
effect if judges routinely disregard it. A meaningful evaluation must
also account for the fact that human judges and AI decision aids may
operate on different information and process that information in
different ways: judges may draw on case-specific cues unavailable to an
algorithm, while algorithmic tools are typically restricted to recorded
inputs and may apply them more consistently.

Our second contribution is to use this framework to synthesize the
empirical evidence across the relevant disciplines. Rather than treating
research on algorithms, human judges, and human-machine interaction as
separate bodies of scholarship, we examine what each contributes to
evaluating AI-assisted judicial decision making and where the links
between them remain underdeveloped.

Our review yields three broad conclusions. First, existing criminal
justice risk assessment instruments possess meaningful predictive power,
but the evidence on their performance remains more limited than is
sometimes assumed. Validation studies vary considerably in design and
reported metrics, often give insufficient attention to the processes
generating observed outcomes, and provide limited evidence about how
particular instruments and modelling approaches perform across contexts.
Second, comparisons between algorithmic and human prediction are
methodologically difficult and yield mixed results. Experimental studies
create comparable prediction tasks but simplify the informational
environment faced by real judges, while observational studies offer
greater realism at the cost of difficult counterfactual and
identification problems. Third, the available evidence suggests that
introducing algorithmic risk assessments into actual judicial decision
making has, so far, produced at most modest effects on pretrial and
sentencing decisions. High rates of judicial override and the limited
evidence on when such overrides improve or worsen decisions point to an
important agenda for future research.

Our discussion focuses on predictive, data-driven risk assessment in
criminal justice. This represents only one class of AI applications in
law. Machine learning, Natural Language Processing and computational
methods are also used or proposed for case triage, information
retrieval, legal document summarization, transcription, evidence
evaluation, and document assembly, among other tasks. We nevertheless
use risk assessment as our focal case because it provides an unusually
developed empirical setting in which to examine the relationship between
prediction and decision making. Outcomes such as failure to appear or
subsequent offending provide comparatively clear benchmarks for
evaluating predictions, making it possible to study questions that are
much harder to address where the relevant outcome -- such as the legal
correctness of an opinion or the appropriateness of a damages award --
is intrinsically difficult to measure.

We conduct a synthetic rather than a systematic review or meta-analysis.
Its purpose is to connect and synthesize empirical research drawn from
several fields -- including computer science, economics, criminology,
and psychology. Relevant studies were identified through targeted
searches using Google's \emph{Scholar Labs}, citation chaining from
influential and recent contributions, and the authors' knowledge of the
respective literatures. We focus on empirical studies of predictive risk
assessment used, or proposed for use, in pretrial, sentencing, and
parole decisions, together with research on human judicial judgment and
human responses to algorithmic advice that bears directly on the three
questions developed in our framework. We exclude work concerned
primarily with other applications of AI to legal tasks, such as legal
information retrieval, document generation, case classification, or
adjudication by fully automated systems, as well as purely doctrinal or
normative discussions except where they clarify the interpretation of
the empirical evidence. The review primarily covers studies in English
published during the period 2010-2025. Our discussion of predictive
validity and impact draws heavily on systematic reviews and
meta-analyses conducted in these specific areas \citep{Desmarais2021z7236,Fazel2022z7093,Lawson2024z7263}.
Because the empirical literature on algorithmic risk assessment is
overwhelmingly concentrated in the United States, our review has a
predominantly US jurisdictional focus. We list the empirical studies
that served as basis for our synthetic review in the Appendix. Our
synthetic approach allows us to integrate and discuss bodies of research
that differ substantially in terminology, methodology, and disciplinary
orientation, but it also entails limitations. In particular, because we
did not conduct a preregistered or exhaustive systematic search, the set
of studies discussed may reflect selection effects arising from database
coverage, citation networks, and our own disciplinary knowledge. Our
conclusions should therefore be understood as a critical synthesis of
the principal empirical findings and methodological issues in the
existing literature, rather than as estimates derived from a formally
exhaustive evidence base.

The paper is organised as follows. We first introduce our analytical
framework. We then discuss machine prediction and field validation
studies. Next we examine the AI-versus-Human literature. Finally, we
consider the AI-plus-Human literature. The concluding Section summarizes
the strengths and limitations of existing work and anticipates future
questions.

\section{Analytical Framework}

Figure~\ref{fig:info-prediction-decision} provides a simple framework for thinking about the role of AI
in judicial decision making.

Criminal risk assessment is about predicting \(Y\), that is: rearrest,
failure to appear before court or new conviction.

Both judges and algorithmic tools form predictions on the basis of
available information, but they do not necessarily rely on the same
information or process it in the same way. An AI system generates a
prediction from observed and recorded characteristics \(X\), such as
age, charges, or prior criminal history. Human judges can draw on the
same information, but may also take into account case-specific factors
that are not captured in the data, such as expressions of remorse or a
gang tattoo. We call these factors \emph{unobservables}, \(U\), because
they are not observed by the algorithm. At the same time, human judgment
may be influenced by noise \(\varepsilon\): salient but weakly
informative or irrelevant cues, including demeanour, fatigue, or other
extraneous factors. We denote\( {\widehat{Y}}_{M} = f(X)\) the
predictions generated by an automated risk assessment system on the
basis of \(X\); and \({\widehat{Y}}_{H} = g(X,U,\varepsilon)\) the
predictions made by human judges. Where \(f\) and \(g\) represent the
predictive function -- mapping information into predictions -- of,
respectively, automated risk assessments instruments and human judges.

In practice, AI does not replace human judgment. Whether AI-aided or
unaided, a human adjudicators make the final decision \(D\) to impose a
monetary bail, to release the convict or to send the defend to prison.

% User-supplied figure, with pasted HTML/Markdown escaping normalized.
\begin{figure}[t]
\centering
\resizebox{\linewidth}{!}{%
\begin{tikzpicture}[
    font=\small,
    >=Latex,
    node distance=0.9cm and 1.2cm,
    info/.style={draw,rounded corners,align=left,text width=4.2cm,
        minimum height=1.0cm,inner sep=4pt},
    proc/.style={draw,rounded corners,align=center,text width=3.1cm,
        minimum height=1.0cm,inner sep=4pt,fill=gray!10},
    stage/.style={font=\bfseries,align=center},
    arr/.style={->,thick},
    noise/.style={->,thick,dotted}
]
\node[stage] (s1) at (0,3.1) {Information environment};
\node[stage] (s2) at (5.7,3.1) {Prediction};
\node[stage] (s3) at (10.0,3.1) {Decision};
\node[info] (x) at (0,1.8)
    {\textbf{Observed information} $(X)$\\
    \scriptsize structured, recorded features\\
    \scriptsize e.g. age, charges, prior record};
\node[info,below=of x] (u)
    {\textbf{Unobservables} $(U)$\\
    \scriptsize relevant case-specific factors not encoded in the data\\
    \scriptsize e.g. remorse, context, gang tattoo};
\node[info,below=of u] (e)
    {\textbf{Noise} $(\varepsilon)$\\
    \scriptsize salient but weakly informative or irrelevant cues\\
    \scriptsize e.g. demeanour, race, fatigue};
\node[proc] (ai) at (5.7,1.8)
    {\textbf{Machine prediction}\\
    \scriptsize $\hat{Y}_{M}=f(X)$};
\node[proc] (human) at (5.7,-0.6)
    {\textbf{Human prediction}\\
    \scriptsize $\hat{Y}_{H}=g(X,U,\varepsilon)$};
\node[proc,text width=3.4cm] (decisionHM) at (10.0,1.45)
    {\textbf{Machine-supported decision}\\
    \scriptsize $D_{H+M}$};
\node[proc,text width=3.4cm] (decisionH) at (10.0,-0.9)
    {\textbf{Human-alone decision}\\
    \scriptsize $D_H$};
\draw[arr] (x.east) -- (ai.west);
\draw[arr] (x.east) -- (human.west);
\draw[arr] (u.east) -- (human.west);
\draw[noise] (e.east) -- (human.west);
\draw[arr] (human.east) -- (decisionH.west);
\draw[arr] (human.east) -- (decisionHM.west);
\draw[arr] (ai.east) -- (decisionHM.west);
\node[align=center,font=\scriptsize] at (5.7,0.75)
    {AI uses only observed inputs};
\node[align=center,font=\scriptsize] at (5.7,-1.75)
    {Humans may use extra signal,\\but may also pick up noise};
\end{tikzpicture}%
}
\caption{Conceptual framework to evaluate Machine, Man + Machine and Man vs Machine research.}
\label{fig:info-prediction-decision}
\end{figure}

These distinctions help clarify the empirical questions that one need to
answer in order to assess how algorithmic decision support systems
affect judging.

Predictive-validity research asks how well the machine predictions
perform (how close \({\widehat{Y}}_{M}\) stands to \(Y\)).
AI-versus-Human research compares machine and human predictions. The
introduction of algorithmic aids in criminal courts and parole boards
normally rests on the belief that machine predictions will be closer to
the true recidivism rate:
\(Y - {\widehat{Y}}_{M} < Y - {\widehat{Y}}_{H}\). AI-plus-Human
research examines what happens when algorithmic prediction is introduced
as a recommendation into the judge's decision process, \(D_{H + M}\).

In order to understand how and why human and machine predictions differ
with respect to accuracy, consistency and fairness, one must consider
not only the informational inputs (\(X\) versus \(X,U,\varepsilon\)) but
also \(f\) and \(g\), how humans and algorithms differently process this
information to arrive at predictions.

Estimating the impact of algorithmic recommendations requires a
comparison between decision making by human alone, \(D_{H}\), and
decision making aided by AI, \(D_{H + M}\), knowing that, at the
decision stage, judges may also weigh considerations other than
\({\widehat{Y}}_{M}\) and \({\widehat{Y}}_{H}\) (e.g. deterrence,
retribution).

\section{Machine Prediction: Field Validation Studies}

This Section discusses the literature assessing the predictive validity
of automated risk assessment instruments.

\subsection{Risk Assessment Algorithms}

Both ``AI'' and ``algorithm'' are capacious terms, that may obscure how
risk assessment instruments actually work.

In some contexts, they refer to symbolic AI methods and expert systems based based on
explicit representations of knowledge, rules, and logical relationships
\citep{Ashley2017z7081}. In the context of risk assessment, however, these
terms are more likely to refer to machine learning methods. Machine learning
encompasses a broad family of techniques based on statistical learning and
optimization, including tree-based methods (CART, random forest,
gradient boosted trees), neural networks and the transformer
architecture underpinning Large Language Models (LLMs). In short,
training a machine learning model to predict \(Y\) from \(X\) involves
fitting many versions of \(f\) to the training data and then selecting
the one that minimizes the prediction error \(Y - {\widehat{Y}}_{M}\).

Despite the word AI being widely used and despite the prominence of
COMPAS in broader discussions of AI and algorithmic fairness
\citep{Hellman2025z7035}, most risk assessment instruments currently deployed
around the world appear to be built around old, actuarial, classical
statistical modelling techniques rather than machine learning methods
\citep{Desmarais2021z7236}.

Equivant (formerly Northpointe), the private, for-profit corporation
delivering COMPAS, provides limited documentation on its inner workings
\citep{Rudin2020z7031}. But studies attempting to reconstruct its methodology
indicate a linear structure approximating a linear additive regression
model \citep{Dressel2018z7032}. Similarly, the weights assigned to the nine
factors used to construct the PSA risk score appear to have been
estimated using logistic regression, with odds ratios rounded to the
nearest integer to ease scoring.\footnote{See Risk Assessment Fact Sheet
  on the Stanford Law School website:
  \url{https://law.stanford.edu/pretrial-risk-assessment-tools-factsheet-project/}.
  (Accessed 18 December 2025.)}

Both PSA and COMPAS bear similarities to the Level of Service-Revised
(LSI-R), one of the most widely used risk assessment instruments
worldwide for sentencing and probation \citep{StephenWormith2018z7224}. Introduced in
the 1990s, LSI-R builds on decades of research in clinical psychology
and criminology. Importantly, it was developed outside the machine
learning paradigm and, instead, presented as a systematic, actuarial
alternative to clinical assessment \citep{Fazel2012z7230,Kehl2017z7228}. In the actuarial or
clinical paradigm, the functional form of \(f\) is not learned from the
data but specified by the researcher on the basis of theory or prior
research.

Real-world applications of advanced, supervised machine learning in the
field of crime risk assessment seem to be rare. One illustration,
though, is the decision aid system used by the Pennsylvania Board of
Probation and Parole described in \citet{Berk2017z7258} and based on random
forest \citep{Breiman2001z7265}.

Because machine learning techniques have been developed with
out-of-sample prediction as their primary goal, they are generally
presumed to outperform simple linear models. Yet real-world applications of advanced, supervised machine learning appear to be rare in the field of crime risk assessment. The one, single instance we could identify is a decision aid system used by the Pennsylvania Board of
Probation and Parole described in \citet{Berk2017z7258} and based on random
forest \citep{Breiman2001z7265}.

From existing
simulations, the evidence that machine learning outperforms older methods used in predicting crime is mixed \citep{Duwe2017z7262,Ghasemi2021z7260,Jung2017z7281,Kleinberg2018z5972,Rudin2020z7031}. One possible
explanation is that gains from machine learning methods depend on the
presence of nonlinear patterns in the data. Machine learning methods
deliver the largest gains for tasks involving high-dimensional data,
such as audio, video or textual data. High-dimensional data are more
likely to contain non-linear patterns. In criminal risk assessment, by
contrast, \(X\) -- the set of predictors -- is typically low
dimensional: age, crime history, pending charges, employment status, vocational background.

Compared to actuarial methods, machine learning models are also more
opaque, which may be a contributing factor to their low diffusion
\citep{Stevenson2024z7208}. As we shall see, though, machine learning models are
often used as benchmark in AI-versus-Machine research, where decisions
by real-world judges are compared to simulated releases based on
AI-generated recommendations. Kleinberg et. al. (2018), for example,
train a gradient boosted trees algorithm for that purpose. The algorithm
builds multiple decision trees sequentially, where each new tree focuses
on correcting the errors made by the previous trees, resulting in a
complex model where the final prediction is essentially the average of
its multiple trees.

\subsection{Predictive Performance}

Even with good data, prediction systems are never error-free. They
produce mispredictions: false positives and false negatives -- the
difference between actual \(Y\) and predicted \(\widehat{Y}\).

The simplest and most straightforward measure of predictive performance
is accuracy: the proportion of correct predictions. However, accuracy is
a coarse and often misleading performance metric. For example, if less
than 5\% of arrestees do actually recidivate, a model that would never
predict reoffence would actually be accurate more than 95\% of the time.

Moreover, the raw output of a predictive model is a probability -- e.g.
a 65\% probability that a given arrestee will reoffend. But for
deployment and evaluation purposes, this numerical value must be
converted into discrete categories: e.g. high/low risk.\footnote{For an
  accessible, thorough discussion of this issue see \citet{Lagioia2023z7157}.}

Determining the threshold for assignment to a category is not an
innocuous question. Should a 45\% chance of recidivism be considered a
predicted positive? A 20\% risk? Should a 35\% risk of recidivism be
classified as high risk? Should the same threshold apply for all
categories of crimes or should a lower threshold apply for violent
crimes? The choice of threshold directly impacts true and false positive
rates. A lower threshold will result in a higher false positive rate. A
higher threshold will result in fewer false positives but will likely
miss more true positives. For real-world applications, the selection of
a specific threshold is a normative rather than an empirical matter (see
discussion of the leniency/incarceration trade-off in Section~\ref{sec:human-judges}).

For that reason, a widely used metric to assess and compare the
performance of predictive models is the area under the curve (AUC), also
known as Receiver Operating Characteristic (ROC). Unlike other
performance metrics, such as precision and specificity, it does not
assume a given threshold, but, instead, measures how well a model can
distinguish between true and false positives across all possible
thresholds. An AUC score of 0.5 indicates performance no better than a
random guess whereas an AUC score of 1 reflects perfect prediction
across all thresholds.

A systematic review of research articles assessing the predictive
validity of six pretrial risk assessment instruments, including PSA and
COMPAS, found AUC values ranging from 0.64 to 0.73 \citep{Desmarais2021z7236}.
Another systematic review looked at nine distinct risk assessment
instruments used in sentencing. It found AUC scores ranging from AUC =
0.42 to AUC = 0.82, and AUC = 0.57 to AUC = 0.75 for studies with sample
size N \textgreater{} 500 \citep{Fazel2022z7093}. An older meta-analysis
compiled estimates for nine risk assessment instruments across 68
independent studies and thirteen countries \citep{Fazel2012z7230}. It reported
an average AUC = 0.66 (CI{[}0.58,0.67{]}) for criminal offending, AUC =
0.72 (CI{[}0.68,0.78{]}) for violent offending and AUC = 0.74
(CI{[}0.66,0.77{]}) for sexual offending.

Another important, threshold-independent performance metric is
calibration, which measures how well expected probabilities compare to
true outcome frequencies. If a model is well-calibrated, released
convicts for which the model predicts a 10\% probability of recidivism
should recidivate, on average, close to 10\% of the time. Unfortunately,
very few predictive validation studies report calibration metrics such
as Expected Calibration Error \citep{Fazel2022z7093}.

\subsection{Prediction Performance Across Groups: Fairness}

Some misprediction is unavoidable. But do error rates vary across
sensitive categories, such as gender and ethnicity?

Biases in data-driven systems may stem from biases embedded in the
historic training data, measurement error or data with underrepresented
groups or categories. Historical criminal justice data reflect
pre-existing societal and institutional biases. Remedying these sources
of bias is rendered difficult by correlations between seemingly
innocuous characteristics -- e.g. ZIP code, age -- and sensitive,
protected traits such as ethnicity, gender or sexual orientation. Risk
assessment instruments do not use race, gender or sexual orientation as
predictor. However, \(X\) typically includes age, which has been shown
to correlate with race in US criminal data, as black defendants tend to
be younger than white defendants \citep{Engel2025z7013,Rudin2020z7031}.

Conversely, it has been shown that ignoring gender can be detrimental to
women, as treating female convicts like male convicts may result in
overprediction of women's likelihood of recidivism \citep{Skeem2016z7238}

There exist various strategies to address biases in data-driven AI
applications. One is via data curation and preprocessing to remove
correlations with sensitive characteristics \citep{Kamiran2012z7042}. Another is
by adding constraints to the optimization objective at the training
stage \citep{Berk2021z7446,Kleinberg2018z5972,Zafar2017z7040}. Yet another is to correct results at the
postprocessing stage \citep{Hardt2016z7037}.

The controversy surrounding COMPAS has sparked the development of formal
measures of fairness against which models can be trained and
benchmarked. These efforts have shown that, except in highly unusual
circumstances where groups exhibit the base rate (e.g. for recidivism),
a model cannot simultaneously satisfy every measure of fairness
\citep{Hellman2025z7035,Lagioia2023z7157}. For example, it is mathematically impossible for a
model simultaneously to satisfy predictive parity (those predicted to
recidivate do so with the same frequency across groups) and equalized
odds (same true positive and false positive rates across groups).

Reviewing 11 studies, \citet{Desmarais2021z7236} found the predictive validity of
pretrial assessment tools to be generally comparable across racial and
ethnic subgroups, but with three studies reporting lower predictive
validity for defendants of color. This assessment, however, is based on
predictive parity -- the prevalent measure of fairness in the predictive
validation literature. Less commonly reported fairness metrics, such as
equalized odds, might well have led to a different assessment
\citep{Berk2021z7446}.

\subsection{Summary}

Existing or proposed risk assessment instruments do possess predictive
power. This achievement aside, though, the literature covered in this
Section reveals significant limitations. Discussions of fairness lack
precision \citep{Berk2021z7446}. Field studies rarely discuss potential
distortions arising from the data generation process, such as sampling
bias \citep{Goel2021z7291}. This is despite the fact that poor, biased data
can distort model evaluation in the same way that it affects model
training. Evaluating the accuracy of a risk assessment instrument
requires a valid measure of \(Y\). In predictive validity studies, \(Y\)
is often measured as re-arrest or as new conviction \citep{Desmarais2021z7236,Fazel2022z7093}.
Yet even with full access to court and police records, many crimes
remain unreported. The discrepancy between recorded crimes and the real
\(Y\) can be further amplified by local and/or time-varying factors,
such as the resources allocated to law enforcement.

The difficulty involved in measuring reoffence entails that one should
have greater confidence in the findings of studies relying on outcomes
that are more likely to be accurate and measured independently of law
enforcement resources or geographic factors such as failure-to-appear or
failure to comply with parole conditions. Overall, though, the available
empirical evidence is not solid enough to establish with confidence
which instruments offer the most reliable predictions or how they would
perform outside the context and data of the particular study
\citep{Fazel2022z7093}.

\section{Human-versus-Machine}

This Section discusses research comparing human and automated risk
assessments.

\subsection{Research on Human Judges}\label{sec:human-judges}

While it is far from furnishing a direct estimate of human prediction
error, the sizable body of research on jurors and professional judges in
psychology, economics, law, and political science provides a first
baseline to understand how AI tools might compare to unaided human
judges.

\subsubsection{Cognitive Biases}

Controlled vignette experiments have reported evidence that jurors and
professional magistrates are susceptible to the following cognitive
biases:

\begin{itemize}
\item
  Confirmation bias \citep{Kukucka2014z6166,OBrien2009z6161,Wistrich2012z7051}.
\item
  Anchoring bias and contrast effect \citep{Bystranowski2021z6189,Englich2006z7060,Rachlinski2012z7062,Rachlinski2015z7059}.
\item
  Framing effects \citep{Guthrie2000z1980,Rachlinski2017z6997}.
\item
  Hindsight bias \citep{Casper1989z7111,Fessel2011z6192,Guthrie2000z1980,Harley2007z7110}.
\item
  Gender bias: benevolent sexism \citep{Rachlinski2021z7112}.
\item
  Base-rate fallacy \citep{Guthrie2000z1980}.
\item
  Out-group bias \citep{Levinson2017z7104,Rachlinski2008z7153,Rachlinski2015z7059}.
\end{itemize}

Observational, archival studies exploiting random or quasi-random
case/juror assignment, complement these findings. They report evidence
consistent with the existence of out-group bias \citep{Choi2022z7108,Ash2025z7106,Abrams2012z7139},
benevolent sexism \citep{Bontrager2013z7138,Philippe2020z7046,Starr2015z7148} and anchoring effects
\citep{Bushway2012z7118}. Studies of asylum and immigration cases have revealed
extreme variations among judges despite random case assignment
\citep{Hangartner2025z7133,RamjiNogales2007z7135}, while sequential negative auto-correlation in decision
patterns suggests that judges occasionally commit the gambler's fallacy
\citep{Chen2016z1987}.

\subsubsection{Simplified Judgment Heuristics}\label{sec:heuristics}

A more limited body of studies indicate that, when required to assess
multiple factors or sources of information, judges often adopt
simplified cognitive strategies. Although judges claim to consider a
wide array of factors (including pending charges, criminal history,
employment, local ties) when setting bail, they appear to rely almost
exclusively on prosecutorial recommendations \citep{Ebbesen1975z7416}.
\citet{Dhami2001z7413} show that the bail decisions of judges in their
experimental setup are best predicted by a simple, noncompensatory
heuristic rule. A heuristic rule using only two factors provided the
best fit for 75.5\% of participants. \citet{Dhami2003z7418} replicated this
finding using observational data from two British courts. Outside
criminal law, \citet{Beebe2006z7420} finds that although US federal courts
claim to apply multifactor tests incorporating up to thirteen factors
when assessing consumer confusion in trademark infringement cases judges
appear to use only two (similarity of the marks and proximity of the
goods). Similar patterns have been found in US copyright fair use
disputes \citep{Beebe2007z7426,Beebe2020z7424} and regarding awards of attorney's fees in
class actions \citep{Eisenberg2004z7421}, where judges claim to weigh multiple
factors, most of which do not have any effects on outcomes.

\subsubsection{Ideological Attitudes}

Empirical research has demonstrated that human judges can be heavily
influenced by their ideological attitudes \citep{Bonica2021z7125,Cohen2019z7136,Dyevre2026z7131,Epstein2013z3184,Hangartner2025z7133,Pinello1999z7129,RamjiNogales2007z7135,Sunstein2007z6847}. When induced
by judicial ideology, variations in sentencing severity, bail or parole
decisions, are not necessarily synonymous with bias in the sense of
departure from rationality. Indeed, such decisions present judges with
an unavoidable trade-off. Jailing defendants tends to reduce crime;
releasing more defendants to increase crime. Ideological differences
over leniency and severity can thus be interpreted as preference for one
type of error over another: less lenient, typically conservative judges
prioritize a low false negative rate (ensuring they don't release people
who might commit crimes) at the cost of a higher false positive rate
(jailing people who won't commit crimes); more lenient, progressive
judges prioritise a lower false positive rate at the cost of a higher
false negative rate. Both experimental and observational evidence,
though, shows that ideology can trigger motivated reasoning, distorting
the evaluation of evidence or assessments regarding the scope and
relevance of precedents \citep{Braman2007z5836,Lu2025z7178}.

\subsubsection{Intuitive Reasoning and Informational Noise}

Judges operate in a noisy environment, where briefs, pleadings, reports,
along with defendants, prosecutors and attorneys, expose them to a
multitude of signals and cues. While it does not specifically address
prediction, the foregoing findings suggest that human judges may indeed
form incorrect predictions because they do not solely consider
informational inputs (\(X,U\)) but also pick irrelevant signals, noise
(\(\varepsilon\)).

Routine reliance on intuitive reasoning suggests that forecasting errors
may also result from \(g,\) the judges' prediction function. ``Fast and
frugal heuristics'' do not always result in bad decisions and can prove
surprisingly accurate \citep{Gigerenzer2000z7172}. But occasional biases and errors
are the flipside of relying on heuristics \citep{Kahneman2011z2342,Rachlinski2017z6997}.

\subsection{Direct Human-versus-Machine Comparisons}

\subsubsection{Methodological Approaches}

Much of the research directly comparing automated risk assessment tools
to human decision makers consists of lab experiments with lay human
participants recruited via crowdsourcing platforms \citep{Biswas2020z7269,Dressel2018z7032,Lin2020z7165}.
Human participants are presented with a vignette including the same
information (age, criminal history) fed to the algorithm and receive a
monetary reward reflecting the quality of their predictions (as measured
by Brier scores).

The other approach is to compare the decisions of real-world judges to
simulations. Observational data, however, come with serious
methodological challenges. The first is the \emph{omitted payoff}
problem. In the field, \({\widehat{Y}}_{H}\) cannot be observed, at
least not directly. Nor can \({\widehat{Y}}_{H}\) easily be inferred
from \(D_{H}\). Indeed, the decision to release or to jail a convict
often involves other considerations in addition to the risk of
recidivism. Sentencing decisions, for example, ordinarily consider just
retribution and deterrence given the gravity of the crime. Without
accounting for these unobserved considerations, assuming that \(D_{H}\)
is a correct proxy for \({\widehat{Y}}_{H}\) is wrong.

Another challenge is known as the \emph{selective label} problem. Crime
outcomes can only be observed for released convicts and defendants.
Whether jailed defendants would have committed a crime had they been
released, by definition, cannot be observed. It might be tempting to
address this counterfactual by imputing jailed defendants the crime risk
of released defendants with the same observed information (\(X\)). This
strategy, however, ignores the role of unobservables, \(U\). If judges
also jail on the basis of \(U\), jailed and released defendants cannot
constitute random samples of the same defendant population
\citep{Lakkaraju2017z7267}.

Some studies do not seek to address these problems but admit them as a
limitation of the reported findings. To simulate recidivism under the
diversion recommendations of a risk assessment instrument for sentencing
introduced in Virginia in 2003, \citet{Stevenson2024z7208} impute detainees the
recidivism rate of released individuals with the same risk score.

\citet{Kleinberg2018z5972}, on the other hand, address these problems head on
through careful sampling, data curation and analysis. First, they focus
on pretrial bail decisions. Judges in their sample are required by law
to consider only the risk of flight or rearrest. This narrow scope
reduces the size of the omitted payoff problem. Second, to impute
post-release behaviour to jailed defendants, they exploit the fact that
cases in their data are randomly assigned to judges who differ in
leniency. Then, for defendants that even the most lenient judges decide
to incarcerate, multiple release practices are simulated to reflect
possible assumptions about the role of unobservables (\(U\)) in
determining incarceration.

\subsubsection{Empirical Findings}

Lab studies report mixed findings. With regard to both predictive
accuracy \citep{Dressel2018z7032,Lin2020z7165} and fairness \citep{Biswas2020z7269,Dressel2018z7032} human
predictions sometimes match, sometimes underperform algorithmic
predictions (with COMPAS or an additive, logistic regression model
commonly offered as AI benchmark). The relatively good performance of
lay human forecasters in \citet{Dressel2018z7032}, though, is likely an artefact
of study design. Participants received immediate feedback on the
accuracy of their predictions, creating an unrealistic decision making
environment. Studies that did not provide immediate feedback reported
considerably poorer human accuracy \citep{Biswas2020z7269,Lin2020z7165}. Another limitation
of existing lab studies is that human participants are only treated with
informational inputs that are consistent across cases and relevant to
risk assessment. As \citet{Goel2021z7291} point out, such ``boosted'' human
predictions are far removed from unaided human judgment in real-world
conditions, where judges are typically presented with a much noisier set
of inputs, including police reports, victim impact statements and
non-structured clinical assessments.

Observational studies, too, report mixed findings. \citet{Kleinberg2018z5972}
compare their well-calibrated algorithm to the judges' release decisions
and their impact on the actual crime rate (primarily measured as failure
to appear). Relating the judges' release rates to the AI-predicted crime
rate, they find an interesting pattern. The curve relating the two rates
is non-linear and flattens out as predicted risk increases. Judges
release defendants with a 75\% crime risk at nearly the same rate as
defendants with a 48\% crime risk. Stricter judges do not prioritize
high risk defendants. Nor do they follow a consistent risk threshold.
Instead, stricter judges jail additional defendants drawn from
throughout the predicted crime risk distribution. This suggests that
automated risk assessment tools may improve on human decisions through
greater consistency. Simulating outcomes under the release
recommendations of their machine learning algorithm, the authors
demonstrate potential of up to 24.7\% reduction in the crime rate, and
up to 41.9\% in the incarceration rate. Moreover, the algorithm can
achieve these simulated gains while simultaneously reducing racial
disparities. These gains are also robust to accounting for potential
omitted pay-off bias associated with the employment or marital status of
defendants.

This study is equally interesting for what it says about human judges.
Compared to the algorithm, human judges especially struggle with
high-risk defendants. Their errors correlate with current charges: they
are most likely to release high-risk people if their current charge is
minor, such as a misdemeanor, but treat low-risk people as if they were
high-risk when their current charge is more serious. Interestingly, a
model trained to predict the judges' decisions, that is \(D_{H}\) rather
than \(Y\), does better at reducing crime than the judges themselves.
Consistent with the empirical research on human judges and jurors
discussed above, these findings suggest that human judges give excessive
weight to highly salient information and often pick noisy signals
\citep{Kleinberg2018z5972}. \citet{Sunstein2019z7444} speculates that the excessive weight
given to current charges may stem from the availability bias.

Using the same dataset as Kleinberg et al. but relaxing assumptions
about the judges' unobserved utility function, \citet{Rambachan2024z7275} found
that replacing judges where systematic mistakes occur with the decision
rule of an ensemble model (averaging an elastic net model and a random
forest model) dominates decision making by humans alone. Results,
however, are more ambiguous (i.e. depend on assumptions about policy
preferences) when all human decisions are replaced by the algorithm's
recommendations. Using a structural model, Rambachan estimates that at
least 20\% of judges make systematic predictive mistakes about
misconduct risk.

To isolate \({\widehat{Y}}_{H}\), \citet{BenMichael2025z7163} consider a setting
where pretrial judges can only impose a signature bond (release on own
recognizance), a small cash bond or a large cash bond. Comparing
human-alone to PSA recommendations, they find the latter to have higher
misclassification and false positive rates for recidivism than human
judges, suggesting that the algorithm is unnecessarily harsher in
imposing cash bails.

From their more tentative simulation, \citet{Stevenson2024z7208} suggest that
systematically following the diversion recommendations of the algorithm
introduced in Virginia would have led to shorter sentence length and
lower incarceration rates at the price of slightly higher recidivism --
an outcome in line with the main objective (namely, reducing
incarceration rates for low risk individuals) of the legislature when
introducing automated risk assessment.

\subsection{Summary}

Comparing human judges to AI algorithms is no easy feat. Experimental
studies seek to hold prediction tasks (and prediction inputs) constant
across AI and human decision makers, but, in doing so, fail to make the
comparison realistic. Observational studies compare AI to realistic
conditions, but struggle to isolate the quantity of interest for valid
comparison. Moreover, where researchers credibly address these
challenges, empirical results may reflect the intrinsic performance of
the particular risk assessment instrument tool being evaluated (PSA,
advanced machine learning model).

While more research would help compare and evaluate a broader array of
existing tools, the studies discussed in this Section nonetheless
contribute new insights and suggestive evidence into how human judges
grapple with noisy information and struggle to maintain consistency
across cases.

\section{Human-Machine Interactions}

Ideally, when AI is introduced in the courtroom, man and machine ought
to complement each other, leveraging their respective strengths while
mutually correcting their respective blind spots: for AI the inability
to see \(U\); for humans inconsistency in processing both \(X\) and
\(U\), along with the tendency to add noise (\(\varepsilon\)).

Yet, optimal use cannot be guaranteed. Judges may systematically
distrust AI recommendations. Or, on the contrary, judges may be
susceptible to the automation bias, overly relying on automated advice
when other sources of information should also be considered
\citep{Goddard2012z7100,Parasuraman2010z7188}.

\subsection{Methodologies}

Similar to AI-versus-Human research, studies investigating AI-plus-Human
interactions are a mixture of lab experiments and observational
analyses. Lab studies compare the recidivism predictions of participants
treated with AI risk assessment to those of a control group
\citep{Green2019z7194,GrgicHlaca2019z7155}. Observational studies typically examine pre- and
post-implementation outcomes, some using propensity score matching,
regression discontinuity or event study designs \citep{Lawson2024z7263,Lowder2021z7204,Stevenson2024z7208,Viljoen2019z7294}.
Drawing on \citet{Kleinberg2018z5972}, \citet{Angelova2025z7253} investigate instances of
human override by exploiting random case assignment and training machine
learning models to predict the judges' own decisions. Meanwhile,
\citet{BenMichael2025z7163} and \citep{Imai2023z6947} report the results of what is, to
date, the first field randomized controlled trial (RCT) testing the
impact of AI risk assessments on judicial decisions.

\subsection{Empirical Findings}

Lab studies have found a modest effect of algorithmic recommendations on
the predictions of human participants. \citet{GrgicHlaca2019z7155} report that
participants changed their predictions when presented with algorithmic
advice in 7\% of cases. \citet{Green2019z7194} find that participants who
received algorithmic recommendations make more accurate predictions than
the control group while still significantly underperforming the
algorithm. The same study found that, while reducing false positive
rates for both black and white defendants, exposure to algorithmic
advice had disparate racial impact for high risk defendants.

In a systematic review and meta-analysis of 22 observational studies,
\citet{Viljoen2019z7294} report small reductions in incarcerations (aggregated
odds ratio {[}\emph{OR}{]} = 0.63, \emph{p} \textless{} 0.001) and
recidivism rates (\emph{OR} = 0.85, \emph{p} \textless{} 0.02). But with
more than half of the studies evaluated as presenting serious risk of
bias, the authors conclude that the overall strength of this evidence
must be regarded as low. Likewise, \citet{Lawson2024z7263} find little evidence
of disparate impact on race and ethnicity in their systematic review of
21 articles but similarly deplore the scarcity of high quality studies.

Based on a proven and rigorous research design, analyses of the first
RCT report a null result \citep{BenMichael2025z7163,Imai2023z6947}. Judges in the treatment group
follow PSA recommendations 70\% of the time. But false negative and
false positive rates for reoffence show no significant difference with
the control group. These results hold across gender and race.

\citet{BenMichael2025z7163}, \citet{Stevenson2024z7208} and \citet{Angelova2025z7253} all report high
incidence of human override of AI recommendations. Isolating the reasons
that spur judges to do so is a question attracting increasing attention
in recent research. In the sentencing context, \citet{Stevenson2024z7208} found
age, marital status, employment and race to be all robust predictors of
deviations from algorithmic advice. Except for race, all characteristics
correlate with more lenient sentencing. Race correlates with longer
sentences for black convicts, which suggests that deviations may be
induced by human out-group bias. \citet{Angelova2025z7253} found that a subset of
judges outperform algorithmic recommendations. To isolate what could
explain the performance differential, the researchers collect additional
data from the case reports, including information on race, residence
(homelessness, out-of-state address), aggravating circumstances and
recommendation from a pretrial officer. A machine learning model trained
with this additional information better predict the decisions of the
outperforming judges but results in worse predictive performance for the
decisions of the underperforming judges. Outperforming judges are
significantly less likely to release defendants with aggravating
conditions and violent charges against an adult. Underperforming judges
are significantly more likely to jail defendants with an out-of-state
address, despite this factor not being particularly predictive of
pretrial misconduct. While the sample of judges included in the study is
small (62), this provides suggestive evidence that outperforming judges
might be systematically better at picking relevant signals from
additional information.

\subsection{Summary}

Research investigating the impact of decision aid systems in the legal
field is still in its infancy. For supporters and critics of algorithmic
risk assessment, existing human-in-the-loop studies offer both
comforting and disappointing takeaways. First, the impact of risk
assessment instruments appears to be, at most, very modest. The fact
that the first field RCT, reasonably powered (sample size \emph{N} =
1891, with incidence close to 30\% for the main outcome variables),
produced a null result implies that the technology is far from having a
transformative impact on criminal justice. Second, available empirical
evidence does not suggest over-reliance, or judges treating automated
systems as cognitive shortcuts. More longitudinal studies are certainly
needed to assess temporal shifts in human override behaviour. But
\citet{Stevenson2024z7208} report evidence of a decrease rather than an increase
in the impact of automated recommendations over time. This should
assuage fears, notably in legal circles, that the technology is causing
automation bias \citep{Romeo2025z7182,Ruschemeier2024z7176}.

As high-quality studies have only considered the impact of a few risk
assessment tools in a small set of (overwhelmingly US) settings, there
is certainly room for future research to refine, and possibly revise,
these general findings.

Tentative attempts to isolate the effects of individual characteristics
and relevant or irrelevant informational cues on overrides point to
important areas for future research across the fields of judicial
behaviour and AI-plus-Human interactions. According to research on the
automation bias \citep{Goddard2012z7100}, human responses to decision aid
systems are moderated by workload as well as individual cognitive
styles. There are potentially important connections between this
research programme and older work on heuristics reviewed in Section
4.1.2. Cross-individual variations may reflect differences in the
heuristics judges employ to simplify their decision making environment.
Given the respective limitations of lab and real-world settings,
significant advances will necessitate a combination of lab experiments,
survey work (to measure cognitive styles), simulations, observational
analysis and, when possible, field RCTs.

\section{Conclusion}

Using criminal justice risk assessment as a focal point, we have tried
to connect three empirical questions central to the introduction of AI
technologies in the judiciary. After presenting a basic analytical
framework, we canvassed and discussed the corresponding literature,
synthesizing the state of the research on each of these three questions.

These three questions have been likened to the steps new drugs need to
pass to be brought to market \citep{Kleinberg2018z5972}. Evaluating predictive
validity is analogous to Phase 0 and Phase 1 trials where a drug must
demonstrate basic safety. Human-versus-Machine studies are analogous to
Phase 2 trials, where at medium scale drugs are compared to placebo or
current best practices. Finally, Human-plus-Machine studies fulfil the
role of Phase 3 trials to quantify impact on key clinical outcomes.

Drawing this parallel helps summarize the results of our synthetic
review. Substantial progress has been made regarding all three steps.
However, important gaps remain. While the impact of AI decision aid
tools on pretrial and sentencing outcomes currently appears less
impressive than some may have hoped and less damaging than some may have
feared, it is not possible to determine on the basis of the available
empirical evidence whether this outcome results from the particular
technology or if tweaked versions of these particular technology (e.g.
using advanced machine learning) might significantly change the outcome.
Improvement upstream, to assess predictive validity more rigorously,
would help downstream Human-versus-Machine as well as Human-plus-Machine
research. Downstream, interesting new questions have emerged: Why do
judges override AI recommendations? Why do they differ in their ability
to make efficient use of aid systems? Answering these questions promises
to shed light on human adjudicators as well as on AI systems, although,
for now, these lines of research remain in their infancy.

Greater cross-disciplinary integration would help. Predictive validation
studies, for example, would gain from systematically reporting the
fairness metrics developed in computer science \citep{Berk2021z7446}. Work in
economics, notably \citet{Kleinberg2018z5972} and \citet{Rambachan2024z7275}, provide strong
examples of careful data construction for field prediction research in
criminology and other disciplines.

Looking forward, as the research integrates and matures, the technology
may also change. Criminal risk assessment is still dominated by the old
clinical-actuarial paradigm. But machine learning models -- as
\citet{Kleinberg2018z5972} show -- can be constantly retrained, tweaked and
rewired to align with legal norms and principles. Sometimes more
transparently than human decision processes. In theory, LLMs offer the
possibility to leverage unstructured data -- e.g. clinical reports --
for behavioral prediction. But this would also raise new questions about
measurement validity and deployment across the three phases of AI
evaluation.

\bibliographystyle{plainnat}
\bibliography{references}

\begin{thebibliography}{86}
\providecommand{\natexlab}[1]{#1}
\providecommand{\url}[1]{\texttt{#1}}
\expandafter\ifx\csname urlstyle\endcsname\relax
  \providecommand{\doi}[1]{doi: #1}\else
  \providecommand{\doi}{doi: \begingroup \urlstyle{rm}\Url}\fi

\bibitem[Abrams et~al.(2012)Abrams, Bertrand, and
  Mullainathan]{Abrams2012z7139}
David~S. Abrams, Marianne Bertrand, and Sendhil Mullainathan.
\newblock {Do Judges Vary in Their Treatment of Race?}
\newblock \emph{The Journal of Legal Studies}, 41\penalty0 (2):\penalty0
  347--383, 2012.
\newblock \doi{10.1086/666006}.

\bibitem[Angelova et~al.(2025)Angelova, Dobbie, and Yang]{Angelova2025z7253}
Victoria Angelova, Will Dobbie, and Crystal~S Yang.
\newblock {Algorithmic Recommendations and Human Discretion}.
\newblock \emph{The Review of Economic Studies}, page rdaf084, 2025.
\newblock \doi{10.1093/restud/rdaf084}.

\bibitem[Ash et~al.(2025)Ash, Asher, Bhowmick, Bhupatiraju, Chen, Devi,
  Goessmann, Novosad, and Siddiqi]{Ash2025z7106}
Elliott Ash, Sam Asher, Aditi Bhowmick, Sandeep Bhupatiraju, Daniel Chen,
  Tanaya Devi, Christoph Goessmann, Paul Novosad, and Bilal Siddiqi.
\newblock {In-group bias in the Indian judiciary: Evidence from 5 million
  criminal cases}.
\newblock \emph{Review of Economics and Statistics}, pages 1--45, 2025.

\bibitem[Ashley(2017)]{Ashley2017z7081}
Kevin~D. Ashley.
\newblock \emph{{Artificial Intelligence and Legal Analytics: New Tools for Law
  Practice in the Digital Age}}.
\newblock Cambridge University Press, 2017.

\bibitem[Beebe(2006)]{Beebe2006z7420}
Barton Beebe.
\newblock {An empirical study of the multifactor tests for trademark
  infringement}.
\newblock \emph{Calif. L. Rev.}, 94:\penalty0 1581, 2006.

\bibitem[Beebe(2007)]{Beebe2007z7426}
Barton Beebe.
\newblock {An empirical study of US copyright fair use opinions, 1978-2005}.
\newblock \emph{U. Pa. L. Rev.}, 156:\penalty0 549, 2007.

\bibitem[Beebe(2020)]{Beebe2020z7424}
Barton Beebe.
\newblock {An Empirical Study of U.S. Copyright Fair Use Opinions Updated,
  1978-2019}.
\newblock \emph{New York University Journal of Intellectual Property \&
  Entertainment Law (JIPEL)}, 10\penalty0 (1):\penalty0 1--39, 2020.

\bibitem[Ben-Michael et~al.(2025)Ben-Michael, Greiner, Huang, Imai, Jiang, and
  Shin]{BenMichael2025z7163}
Eli Ben-Michael, D.~James Greiner, Melody Huang, Kosuke Imai, Zhichao Jiang,
  and Sooahn Shin.
\newblock {Does AI help humans make better decisions? A statistical evaluation
  framework for experimental and observational studies}.
\newblock \emph{Proceedings of the National Academy of Sciences}, 122\penalty0
  (38):\penalty0 e2505106122, 2025.
\newblock \doi{10.1073/pnas.2505106122}.

\bibitem[Berk(2017)]{Berk2017z7258}
Richard Berk.
\newblock {An impact assessment of machine learning risk forecasts on parole
  board decisions and recidivism}.
\newblock \emph{Journal of Experimental Criminology}, 13\penalty0 (2):\penalty0
  193--216, 2017.
\newblock \doi{10.1007/s11292-017-9286-2}.

\bibitem[Berk et~al.(2021)Berk, Heidari, Jabbari, Kearns, and
  Roth]{Berk2021z7446}
Richard Berk, Hoda Heidari, Shahin Jabbari, Michael Kearns, and Aaron Roth.
\newblock {Fairness in Criminal Justice Risk Assessments: The State of the
  Art}.
\newblock \emph{Sociological Methods \& Research}, 50\penalty0 (1):\penalty0
  3--44, 2021.
\newblock \doi{10.1177/0049124118782533}.

\bibitem[Biswas et~al.(2020)Biswas, Kolczynska, Rantanen, and
  Rozenshtein]{Biswas2020z7269}
Arpita Biswas, Marta Kolczynska, Saana Rantanen, and Polina Rozenshtein.
\newblock {The Role of In-Group Bias and Balanced Data: A Comparison of Human
  and Machine Recidivism Risk Predictions}.
\newblock In \emph{Proceedings of the 3rd ACM SIGCAS Conference on Computing
  and Sustainable Societies}, pages 97--104, New York, NY, USA, 2020.
  Association for Computing Machinery.
\newblock \doi{10.1145/3378393.3402507}.

\bibitem[Bonica and Sen(2021)]{Bonica2021z7125}
Adam Bonica and Maya Sen.
\newblock {Estimating Judicial Ideology}.
\newblock \emph{Journal of Economic Perspectives}, 35\penalty0 (1):\penalty0
  97--118, 2021.
\newblock \doi{10.1257/jep.35.1.97}.

\bibitem[Bontrager et~al.(2013)Bontrager, Barrick, and
  Stupi]{Bontrager2013z7138}
Stephanie Bontrager, Kelle Barrick, and Elizabeth Stupi.
\newblock {Gender and sentencing: A meta-analysis of contemporary research}.
\newblock \emph{J. Gender Race \& Just.}, 16:\penalty0 349, 2013.

\bibitem[Braman and Nelson(2007)]{Braman2007z5836}
Eileen Braman and Thomas~E. Nelson.
\newblock {Mechanism of Motivated Reasoning? Analogical Perception in
  Discrimination Disputes}.
\newblock \emph{American Journal of Political Science}, 51\penalty0
  (4):\penalty0 940--956, 2007.
\newblock \doi{10.1111/j.1540-5907.2007.00290.x}.

\bibitem[Breiman(2001)]{Breiman2001z7265}
Leo Breiman.
\newblock {Random Forests}.
\newblock \emph{Machine Learning}, 45\penalty0 (1):\penalty0 5--32, 2001.
\newblock \doi{10.1023/A:1010933404324}.

\bibitem[Bushway et~al.(2012)Bushway, Owens, and Piehl]{Bushway2012z7118}
Shawn~D. Bushway, Emily~G. Owens, and Anne~Morrison Piehl.
\newblock {Sentencing Guidelines and Judicial Discretion: Quasi-Experimental
  Evidence from Human Calculation Errors}.
\newblock \emph{Journal of Empirical Legal Studies}, 9\penalty0 (2):\penalty0
  291--319, 2012.
\newblock \doi{10.1111/j.1740-1461.2012.01254.x}.

\bibitem[Bystranowski et~al.(2021)Bystranowski, Janik, Próchnicki, and
  Skórska]{Bystranowski2021z6189}
Piotr Bystranowski, Bartosz Janik, Maciej Próchnicki, and Paulina Skórska.
\newblock {Anchoring effect in legal decision-making: A meta-analysis.}
\newblock \emph{Law and Human Behavior}, 45\penalty0 (1):\penalty0 1, 2021.

\bibitem[Casper et~al.(1989)Casper, Benedict, and Perry]{Casper1989z7111}
Jonathan~D. Casper, Kennette Benedict, and Jo~L. Perry.
\newblock {Juror decision making, attitudes, and the hindsight bias}.
\newblock \emph{Law and Human Behavior}, 13\penalty0 (3):\penalty0 291--310,
  1989.
\newblock \doi{10.1007/BF01067031}.

\bibitem[Chen et~al.(2016)Chen, Moskowitz, and Shue]{Chen2016z1987}
Daniel Chen, Tobias~J. Moskowitz, and Kelly Shue.
\newblock {Decision-Making Under the Gambler's Fallacy: Evidence from Asylum
  Judges, Loan Officers, and Baseball Umpires}.
\newblock \emph{The Quarterly Journal of Economics}, page qjw017, 2016.

\bibitem[Choi et~al.(2022)Choi, Harris, and Shen-Bayh]{Choi2022z7108}
Donghyun~Danny Choi, J.~Andrew Harris, and Fiona Shen-Bayh.
\newblock {Ethnic Bias in Judicial Decision Making: Evidence from Criminal
  Appeals in Kenya}.
\newblock \emph{American Political Science Review}, 116\penalty0 (3):\penalty0
  1067--1080, 2022.
\newblock \doi{10.1017/S000305542100143X}.

\bibitem[Cohen and Yang(2019)]{Cohen2019z7136}
Alma Cohen and Crystal~S. Yang.
\newblock {Judicial Politics and Sentencing Decisions}.
\newblock \emph{American Economic Journal: Economic Policy}, 11\penalty0
  (1):\penalty0 160--191, 2019.
\newblock \doi{10.1257/pol.20170329}.

\bibitem[Desmarais et~al.(2021)Desmarais, Zottola, Duhart~Clarke, and
  Lowder]{Desmarais2021z7236}
Sarah~L. Desmarais, Samantha~A. Zottola, Sarah~E. Duhart~Clarke, and Evan~M.
  Lowder.
\newblock {Predictive Validity of Pretrial Risk Assessments: A Systematic
  Review of the Literature}.
\newblock \emph{Criminal Justice and Behavior}, 48\penalty0 (4):\penalty0
  398--420, 2021.
\newblock \doi{10.1177/0093854820932959}.

\bibitem[Dhami(2003)]{Dhami2003z7418}
Mandeep~K. Dhami.
\newblock {Psychological Models of Professional Decision Making}.
\newblock \emph{Psychological Science}, 14\penalty0 (2):\penalty0 175--180,
  2003.
\newblock \doi{10.1111/1467-9280.01438}.

\bibitem[Dhami and Ayton(2001)]{Dhami2001z7413}
Mandeep~K. Dhami and Peter Ayton.
\newblock {Bailing and jailing the fast and frugal way}.
\newblock \emph{Journal of Behavioral Decision Making}, 14\penalty0
  (2):\penalty0 141--168, 2001.
\newblock \doi{10.1002/bdm.371}.

\bibitem[Dressel and Farid(2018)]{Dressel2018z7032}
Julia Dressel and Hany Farid.
\newblock {The accuracy, fairness, and limits of predicting recidivism}.
\newblock \emph{Science Advances}, 4\penalty0 (1):\penalty0 eaao5580, 2018.
\newblock \doi{10.1126/sciadv.aao5580}.

\bibitem[Duwe and Kim(2017)]{Duwe2017z7262}
Grant Duwe and KiDeuk Kim.
\newblock {Out With the Old and in With the New? An Empirical Comparison of
  Supervised Learning Algorithms to Predict Recidivism}.
\newblock \emph{Criminal Justice Policy Review}, 28\penalty0 (6):\penalty0
  570--600, 2017.
\newblock \doi{10.1177/0887403415604899}.

\bibitem[Dyevre and Rodilla~Lazaro(2026)]{Dyevre2026z7131}
Arthur Dyevre and Andreu Rodilla~Lazaro.
\newblock {Judicial Polarization in Parliamentary Democracies: Evidence from
  the Spanish Constitutional Tribunal}.
\newblock \emph{European Journal of Political Research}, 2026.

\bibitem[Ebbesen and Konecni(1975)]{Ebbesen1975z7416}
Ebbe~B. Ebbesen and Vladimir~J. Konecni.
\newblock {Decision making and information integration in the courts: The
  setting of bail.}
\newblock \emph{Journal of Personality and Social Psychology}, 32\penalty0
  (5):\penalty0 805, 1975.

\bibitem[Eisenberg and Miller(2004)]{Eisenberg2004z7421}
Theodore Eisenberg and Geoffrey~P. Miller.
\newblock {Attorney Fees in Class Action Settlements: An Empirical Study}.
\newblock \emph{Journal of Empirical Legal Studies}, 1\penalty0 (1):\penalty0
  27--78, 2004.
\newblock \doi{10.1111/j.1740-1461.2004.00002.x}.

\bibitem[Engel et~al.(2025)Engel, Linhardt, and Schubert]{Engel2025z7013}
Christoph Engel, Lorenz Linhardt, and Marcel Schubert.
\newblock {Code is law: how COMPAS affects the way the judiciary handles the
  risk of recidivism}.
\newblock \emph{Artificial Intelligence and Law}, 33\penalty0 (2):\penalty0
  383--404, 2025.
\newblock \doi{10.1007/s10506-024-09389-8}.

\bibitem[Englich et~al.(2006)Englich, Mussweiler, and Strack]{Englich2006z7060}
Birte Englich, Thomas Mussweiler, and Fritz Strack.
\newblock {Playing Dice With Criminal Sentences: The Influence of Irrelevant
  Anchors on Experts' Judicial Decision Making}.
\newblock \emph{Personality and Social Psychology Bulletin}, 32\penalty0
  (2):\penalty0 188--200, 2006.
\newblock \doi{10.1177/0146167205282152}.

\bibitem[Epstein et~al.(2013)Epstein, Landes, and Posner]{Epstein2013z3184}
Lee Epstein, William~M. Landes, and Richard~A. Posner.
\newblock \emph{{The Behavior of Federal Judges: A Theoretical and Empirical
  Study of Rational Choice}}.
\newblock Harvard University Press, 2013.

\bibitem[Fazel et~al.(2012)Fazel, Singh, Doll, and Grann]{Fazel2012z7230}
Seena Fazel, Jay~P. Singh, Helen Doll, and Martin Grann.
\newblock {Use of risk assessment instruments to predict violence and
  antisocial behaviour in 73 samples involving 24 827 people: systematic review
  and meta-analysis}.
\newblock \emph{BMJ}, 345:\penalty0 e4692, 2012.
\newblock \doi{10.1136/bmj.e4692}.

\bibitem[Fazel et~al.(2022)Fazel, Burghart, Fanshawe, Gil, Monahan, and
  Yu]{Fazel2022z7093}
Seena Fazel, Matthias Burghart, Thomas Fanshawe, Sharon~Danielle Gil, John
  Monahan, and Rongqin Yu.
\newblock {The predictive performance of criminal risk assessment tools used at
  sentencing: Systematic review of validation studies}.
\newblock \emph{Journal of Criminal Justice}, 81:\penalty0 101902, 2022.
\newblock \doi{10.1016/j.jcrimjus.2022.101902}.

\bibitem[Fessel and Roese(2011)]{Fessel2011z6192}
Florian Fessel and Neal~J. Roese.
\newblock {Hindsight Bias, Visual Aids, and Legal Decision Making: Timing is
  Everything}.
\newblock \emph{Social and Personality Psychology Compass}, 5\penalty0
  (4):\penalty0 180--193, 2011.
\newblock \doi{10.1111/j.1751-9004.2011.00343.x}.

\bibitem[Ghasemi et~al.(2021)Ghasemi, Anvari, Atapour, Stephen~wormith,
  Stockdale, and Spiteri]{Ghasemi2021z7260}
Mehdi Ghasemi, Daniel Anvari, Mahshid Atapour, J.~Stephen~wormith, Keira~C.
  Stockdale, and Raymond~J. Spiteri.
\newblock {The Application of Machine Learning to a General Risk--Need
  Assessment Instrument in the Prediction of Criminal Recidivism}.
\newblock \emph{Criminal Justice and Behavior}, 48\penalty0 (4):\penalty0
  518--538, 2021.
\newblock \doi{10.1177/0093854820969753}.

\bibitem[Gigerenzer et~al.(2000)Gigerenzer, Todd, and
  Group]{Gigerenzer2000z7172}
Gerd Gigerenzer, Peter~M. Todd, and ABC~Research Group.
\newblock \emph{{Simple Heuristics that Make Us Smart}}.
\newblock Oxford University Press, 2000.

\bibitem[Goddard et~al.(2012)Goddard, Roudsari, and Wyatt]{Goddard2012z7100}
Kate Goddard, Abdul Roudsari, and Jeremy~C Wyatt.
\newblock {Automation bias: a systematic review of frequency, effect mediators,
  and mitigators}.
\newblock \emph{Journal of the American Medical Informatics Association},
  19\penalty0 (1):\penalty0 121--127, 2012.
\newblock \doi{10.1136/amiajnl-2011-000089}.

\bibitem[Goel et~al.(2021)Goel, Shroff, Skeem, and Slobogin]{Goel2021z7291}
Sharad Goel, Ravi Shroff, Jennifer Skeem, and Christopher Slobogin.
\newblock {The accuracy, equity, and jurisprudence of criminal risk
  assessment}.
\newblock In \emph{Research handbook on big data law}, pages 9--28. Edward
  Elgar Publishing, 2021.
\newblock URL
  \url{https://www.elgaronline.com/abstract/edcoll/9781788972819/9781788972819.00007.xml}.

\bibitem[Green and Chen(2019)]{Green2019z7194}
Ben Green and Yiling Chen.
\newblock {Disparate Interactions: An Algorithm-in-the-Loop Analysis of
  Fairness in Risk Assessments}.
\newblock In \emph{Proceedings of the Conference on Fairness, Accountability,
  and Transparency}, pages 90--99, New York, NY, USA, 2019. Association for
  Computing Machinery.
\newblock \doi{10.1145/3287560.3287563}.

\bibitem[Grgić-Hlača et~al.(2019)Grgić-Hlača, Engel, and
  Gummadi]{GrgicHlaca2019z7155}
Nina Grgić-Hlača, Christoph Engel, and Krishna~P. Gummadi.
\newblock {Human Decision Making with Machine Assistance: An Experiment on
  Bailing and Jailing}.
\newblock \emph{Proceedings of the ACM on Human-Computer Interaction},
  3\penalty0 (CSCW):\penalty0 1--25, 2019.
\newblock \doi{10.1145/3359280}.

\bibitem[Gursoy and Kakadiaris(2022)]{Gursoy2022z7009}
Furkan Gursoy and Ioannis~A. Kakadiaris.
\newblock {Equal confusion fairness: Measuring group-based disparities in
  automated decision systems}.
\newblock In \emph{2022 IEEE International Conference on Data Mining Workshops
  (ICDMW)}, pages 137--146. IEEE, 2022.
\newblock URL
  \url{https://ieeexplore.ieee.org/abstract/document/10029385/?casa_token=NtHLg1nhFTwAAAAA:vYxUoW_wooVMz8yFOfdYoV75ooU-iVpZFqz4HjgLdjn5JUn51ej0i_C9RsnZzEWdN9lKYVtyUwU}.

\bibitem[Guthrie et~al.(2000)Guthrie, Rachlinski, and
  Wistrich]{Guthrie2000z1980}
Chris Guthrie, Jeffrey~J. Rachlinski, and Andrew~J. Wistrich.
\newblock {Inside the judicial mind}.
\newblock \emph{Cornell L. Rev.}, 86:\penalty0 777, 2000.

\bibitem[Hangartner et~al.(2025)Hangartner, Lauderdale, and
  Spirig]{Hangartner2025z7133}
Dominik Hangartner, Benjamin~E. Lauderdale, and Judith Spirig.
\newblock {Inferring individual preferences from group decisions: judicial
  preference variation and aggregation on collegial courts}.
\newblock \emph{British Journal of Political Science}, 55:\penalty0 e163, 2025.

\bibitem[Hardt et~al.(2016)Hardt, Price, and Srebro]{Hardt2016z7037}
Moritz Hardt, Eric Price, and Nathan Srebro.
\newblock {Equality of Opportunity in Supervised Learning}.
\newblock arXiv:1610.02413, 2016.

\bibitem[Harley(2007)]{Harley2007z7110}
Erin~M. Harley.
\newblock {Hindsight Bias In Legal Decision Making}.
\newblock \emph{Social Cognition}, 25\penalty0 (1):\penalty0 48--63, 2007.
\newblock \doi{10.1521/soco.2007.25.1.48}.

\bibitem[Hellman(2025)]{Hellman2025z7035}
Deborah Hellman.
\newblock {Algorithmic Fairness}.
\newblock In Edward~N. Zalta and Uri Nodelman, editors, \emph{The Stanford
  Encyclopedia of Philosophy}. Metaphysics Research Lab, Stanford University,
  fall 2025 edition, 2025.
\newblock URL
  \url{https://plato.stanford.edu/archives/fall2025/entries/algorithmic-fairness/}.

\bibitem[Imai et~al.(2023)Imai, Jiang, Greiner, Halen, and Shin]{Imai2023z6947}
Kosuke Imai, Zhichao Jiang, D~James Greiner, Ryan Halen, and Sooahn Shin.
\newblock {Experimental evaluation of algorithm-assisted human decision-making:
  application to pretrial public safety assessment*}.
\newblock \emph{Journal of the Royal Statistical Society Series A: Statistics
  in Society}, 186\penalty0 (2):\penalty0 167--189, 2023.
\newblock \doi{10.1093/jrsssa/qnad010}.

\bibitem[Jung et~al.(2017)Jung, Concannon, Shroff, Goel, and
  Goldstein]{Jung2017z7281}
Jongbin Jung, Connor Concannon, Ravi Shroff, Sharad Goel, and Daniel~G.
  Goldstein.
\newblock {Simple rules for complex decisions}.
\newblock arXiv:1702.04690, 2017.

\bibitem[Kahneman(2011)]{Kahneman2011z2342}
Daniel Kahneman.
\newblock \emph{{Thinking, Fast and Slow}}.
\newblock Penguin UK, 2011.

\bibitem[Kamiran and Calders(2012)]{Kamiran2012z7042}
Faisal Kamiran and Toon Calders.
\newblock {Data preprocessing techniques for classification without
  discrimination}.
\newblock \emph{Knowledge and Information Systems}, 33\penalty0 (1):\penalty0
  1--33, 2012.
\newblock \doi{10.1007/s10115-011-0463-8}.

\bibitem[Kehl and Kessler(2017)]{Kehl2017z7228}
Danielle Kehl and Sam Kessler.
\newblock {Algorithms in the criminal justice system: assessing the use of risk
  assessments in sentencing}, 2017.
\newblock URL
  \url{https://dash.harvard.edu/entities/publication/73120379-010a-6bd4-e053-0100007fdf3b}.

\bibitem[Kleinberg et~al.(2018)Kleinberg, Lakkaraju, Leskovec, Ludwig, and
  Mullainathan]{Kleinberg2018z5972}
Jon Kleinberg, Himabindu Lakkaraju, Jure Leskovec, Jens Ludwig, and Sendhil
  Mullainathan.
\newblock {Human decisions and machine predictions}.
\newblock \emph{The quarterly journal of economics}, 133\penalty0 (1):\penalty0
  237--293, 2018.

\bibitem[Kukucka and Kassin(2014)]{Kukucka2014z6166}
Jeff Kukucka and Saul~M. Kassin.
\newblock {Do confessions taint perceptions of handwriting evidence? An
  empirical test of the forensic confirmation bias}.
\newblock \emph{Law and Human Behavior}, 38\penalty0 (3):\penalty0 256--270,
  2014.
\newblock \doi{10.1037/lhb0000066}.

\bibitem[Lagioia et~al.(2023)Lagioia, Rovatti, and Sartor]{Lagioia2023z7157}
Francesca Lagioia, Riccardo Rovatti, and Giovanni Sartor.
\newblock {Algorithmic fairness through group parities? The case of
  COMPAS-SAPMOC}.
\newblock \emph{AI \& SOCIETY}, 38\penalty0 (2):\penalty0 459--478, 2023.
\newblock \doi{10.1007/s00146-022-01441-y}.

\bibitem[Lakkaraju et~al.(2017)Lakkaraju, Kleinberg, Leskovec, Ludwig, and
  Mullainathan]{Lakkaraju2017z7267}
Himabindu Lakkaraju, Jon Kleinberg, Jure Leskovec, Jens Ludwig, and Sendhil
  Mullainathan.
\newblock {The Selective Labels Problem: Evaluating Algorithmic Predictions in
  the Presence of Unobservables}.
\newblock In \emph{Proceedings of the 23rd ACM SIGKDD International Conference
  on Knowledge Discovery and Data Mining}, pages 275--284, New York, NY, USA,
  2017. Association for Computing Machinery.
\newblock \doi{10.1145/3097983.3098066}.

\bibitem[Lawson et~al.(2024)Lawson, Narkewicz, and Vincent]{Lawson2024z7263}
Spencer~G. Lawson, Emma~L. Narkewicz, and Gina~M. Vincent.
\newblock {Disparate impact of risk assessment instruments: A systematic
  review}.
\newblock \emph{Law and Human Behavior}, 48\penalty0 (5-6):\penalty0 427--440,
  2024.
\newblock \doi{10.1037/lhb0000582}.

\bibitem[Levinson et~al.(2017)Levinson, Bennett, and Hioki]{Levinson2017z7104}
Justin~D. Levinson, Mark~W. Bennett, and Koichi Hioki.
\newblock {Judging Implicit Bias: A National Empirical Study of Judicial
  Sterotypes}.
\newblock \emph{Florida Law Review}, 69\penalty0 (1):\penalty0 63--114, 2017.

\bibitem[Lin et~al.(2020)Lin, Jung, Goel, and Skeem]{Lin2020z7165}
Zhiyuan~“Jerry” Lin, Jongbin Jung, Sharad Goel, and Jennifer Skeem.
\newblock {The limits of human predictions of recidivism}.
\newblock \emph{Science Advances}, 6\penalty0 (7):\penalty0 eaaz0652, 2020.
\newblock \doi{10.1126/sciadv.aaz0652}.

\bibitem[Lowder et~al.(2021)Lowder, Diaz, Grommon, and Ray]{Lowder2021z7204}
Evan~M. Lowder, Carmen~L. Diaz, Eric Grommon, and Bradley~R. Ray.
\newblock {Effects of pretrial risk assessments on release decisions and
  misconduct outcomes relative to practice as usual}.
\newblock \emph{Journal of Criminal Justice}, 73:\penalty0 101754, 2021.
\newblock \doi{10.1016/j.jcrimjus.2020.101754}.

\bibitem[Lu and Chen(2025)]{Lu2025z7178}
Wei Lu and Daniel~L. Chen.
\newblock {Motivated reasoning in the field: polarization of prose, precedent,
  and policy in US Circuit Courts, 1891--2013}.
\newblock \emph{Plos One}, 20\penalty0 (3):\penalty0 e0318790, 2025.

\bibitem[Niblett(2024)]{Niblett2024z7016}
Anthony Niblett.
\newblock {Artificial Intelligence and Judging}.
\newblock In Lee Epstein, Gunnar Grendstad, Urška Šadl, and Keren Weinshall,
  editors, \emph{The Oxford Handbook of Comparative Judicial Behaviour},
  page~0. Oxford University Press, 2024.
\newblock \doi{10.1093/oxfordhb/9780192898579.013.46}.

\bibitem[O'Brien(2009)]{OBrien2009z6161}
Barbara O'Brien.
\newblock {Prime suspect: An examination of factors that aggravate and
  counteract confirmation bias in criminal investigations}.
\newblock \emph{Psychology, Public Policy, and Law}, 15\penalty0 (4):\penalty0
  315--334, 2009.
\newblock \doi{10.1037/a0017881}.

\bibitem[Parasuraman and Manzey(2010)]{Parasuraman2010z7188}
Raja Parasuraman and Dietrich Manzey.
\newblock {Complacency and Bias in Human Use of Automation: An Attentional
  Integration}.
\newblock \emph{Human factors}, 52:\penalty0 381--410, 2010.
\newblock \doi{10.1177/0018720810376055}.

\bibitem[Philippe(2020)]{Philippe2020z7046}
Arnaud Philippe.
\newblock {Gender Disparities in Sentencing}.
\newblock \emph{Economica}, 87\penalty0 (348):\penalty0 1037--1077, 2020.
\newblock \doi{10.1111/ecca.12333}.

\bibitem[Pinello(1999)]{Pinello1999z7129}
Daniel~R. Pinello.
\newblock {Linking Party to Judicial Ideology in American Courts: A
  Meta-analysis}.
\newblock \emph{The Justice System Journal}, 20\penalty0 (3):\penalty0
  219--254, 1999.

\bibitem[Rachlinski and Wistrich(2017)]{Rachlinski2017z6997}
Jeffrey~J. Rachlinski and Andrew~J. Wistrich.
\newblock {Judging the Judiciary by the Numbers: Empirical Research on Judges}.
\newblock \emph{Annual Review of Law and Social Science}, 13\penalty0
  (1):\penalty0 203--229, 2017.
\newblock \doi{10.1146/annurev-lawsocsci-110615-085032}.

\bibitem[Rachlinski and Wistrich(2021)]{Rachlinski2021z7112}
Jeffrey~J. Rachlinski and Andrew~J. Wistrich.
\newblock {Benevolent sexism in judges}.
\newblock \emph{San Diego L. Rev.}, 58:\penalty0 101, 2021.

\bibitem[Rachlinski et~al.(2008)Rachlinski, Johnson, Wistrich, and
  Guthrie]{Rachlinski2008z7153}
Jeffrey~J. Rachlinski, Sheri~Lynn Johnson, Andrew~J. Wistrich, and Chris
  Guthrie.
\newblock {Does Unconscious Racial Bias Affect Trial Judges}.
\newblock \emph{Notre Dame Law Review}, 84\penalty0 (3):\penalty0 1195--1246,
  2008.

\bibitem[Rachlinski et~al.(2012)Rachlinski, Wistrich, and
  Guthrie]{Rachlinski2012z7062}
Jeffrey~J. Rachlinski, Andrew~J. Wistrich, and Chris Guthrie.
\newblock {Altering attention in adjudication}.
\newblock \emph{UCLA L. Rev.}, 60:\penalty0 1586, 2012.

\bibitem[Rachlinski et~al.(2015)Rachlinski, Wistrich, and
  Guthrie]{Rachlinski2015z7059}
Jeffrey~J. Rachlinski, Andrew~J. Wistrich, and Chris Guthrie.
\newblock {Can judges make reliable numeric judgments: distorted damages and
  skewed sentences}.
\newblock \emph{Ind. LJ}, 90:\penalty0 695, 2015.

\bibitem[Rambachan(2024)]{Rambachan2024z7275}
Ashesh Rambachan.
\newblock {Identifying Prediction Mistakes in Observational Data*}.
\newblock \emph{The Quarterly Journal of Economics}, 139\penalty0 (3):\penalty0
  1665--1711, 2024.
\newblock \doi{10.1093/qje/qjae013}.

\bibitem[Ramji-Nogales et~al.(2007)Ramji-Nogales, Schoenholtz, and
  Schrag]{RamjiNogales2007z7135}
Jaya Ramji-Nogales, Andrew~I. Schoenholtz, and Philip~G. Schrag.
\newblock {Refugee roulette: Disparities in asylum adjudication}.
\newblock \emph{Stan. L. Rev.}, 60:\penalty0 295, 2007.

\bibitem[Romeo and Conti(2025)]{Romeo2025z7182}
Giuseppe Romeo and Daniela Conti.
\newblock {Exploring automation bias in human--AI collaboration: a review and
  implications for explainable AI}.
\newblock \emph{AI \& SOCIETY}, 2025.
\newblock \doi{10.1007/s00146-025-02422-7}.

\bibitem[Rudin et~al.(2020)Rudin, Wang, and Coker]{Rudin2020z7031}
Cynthia Rudin, Caroline Wang, and Beau Coker.
\newblock {The Age of Secrecy and Unfairness in Recidivism Prediction}.
\newblock \emph{Harvard Data Science Review}, 2\penalty0 (1), 2020.
\newblock URL \url{http://hdsr.mitpress.mit.edu/pub/7z10o269/release/3}.

\bibitem[Ruschemeier and Hondrich(2024)]{Ruschemeier2024z7176}
Hannah Ruschemeier and Lukas~J. Hondrich.
\newblock {Automation bias in public administration -- an interdisciplinary
  perspective from law and psychology}.
\newblock \emph{Government Information Quarterly}, 41\penalty0 (3):\penalty0
  101953, 2024.
\newblock \doi{10.1016/j.giq.2024.101953}.

\bibitem[Salo et~al.(2019)Salo, Laaksonen, and Santtila]{Salo2019z7091}
Benny Salo, Toni Laaksonen, and Pekka Santtila.
\newblock {Predictive Power of Dynamic (vs. Static) Risk Factors in the Finnish
  Risk and Needs Assessment Form}.
\newblock \emph{Criminal Justice and Behavior}, 46\penalty0 (7):\penalty0
  939--960, 2019.
\newblock \doi{10.1177/0093854819848793}.

\bibitem[Skeem et~al.(2016)Skeem, Monahan, and Lowenkamp]{Skeem2016z7238}
Jennifer Skeem, John Monahan, and Christopher Lowenkamp.
\newblock {Gender, risk assessment, and sanctioning: The cost of treating women
  like men.}
\newblock \emph{Law and human behavior}, 40\penalty0 (5):\penalty0 580, 2016.

\bibitem[Starr(2015)]{Starr2015z7148}
Sonja~B. Starr.
\newblock {Estimating Gender Disparities in Federal Criminal Cases}.
\newblock \emph{American Law and Economics Review}, 17\penalty0 (1):\penalty0
  127--159, 2015.
\newblock \doi{10.1093/aler/ahu010}.

\bibitem[Stephen~Wormith and Bonta(2018)]{StephenWormith2018z7224}
J.~Stephen~Wormith and James Bonta.
\newblock {The Level of Service (LS) Instruments}.
\newblock In \emph{Handbook of Recidivism Risk/Needs Assessment Tools}, pages
  117--145. John Wiley \& Sons, Ltd, 2018.
\newblock \doi{10.1002/9781119184256.ch6}.

\bibitem[Stevenson and Doleac(2024)]{Stevenson2024z7208}
Megan~T. Stevenson and Jennifer~L. Doleac.
\newblock {Algorithmic Risk Assessment in the Hands of Humans}.
\newblock \emph{American Economic Journal: Economic Policy}, 16\penalty0
  (4):\penalty0 382--414, 2024.
\newblock \doi{10.1257/pol.20220620}.

\bibitem[Sunstein(2019)]{Sunstein2019z7444}
Cass~R. Sunstein.
\newblock {Algorithms, correcting biases}.
\newblock \emph{Social Research: An International Quarterly}, 86\penalty0
  (2):\penalty0 499--511, 2019.

\bibitem[Sunstein et~al.(2007)Sunstein, Schkade, Ellman, and
  Sawicki]{Sunstein2007z6847}
Cass~R. Sunstein, David Schkade, Lisa~M. Ellman, and Andres Sawicki.
\newblock \emph{{Are judges political? An empirical analysis of the federal
  judiciary}}.
\newblock Rowman \& Littlefield, 2007.

\bibitem[Viljoen et~al.(2019)Viljoen, Jonnson, Cochrane, Vargen, and
  Vincent]{Viljoen2019z7294}
Jodi~L. Viljoen, Melissa~R. Jonnson, Dana~M. Cochrane, Lee~M. Vargen, and
  Gina~M. Vincent.
\newblock {Impact of risk assessment instruments on rates of pretrial
  detention, postconviction placements, and release: A systematic review and
  meta-analysis.}
\newblock \emph{Law and Human Behavior}, 43\penalty0 (5):\penalty0 397, 2019.

\bibitem[Wistrich and Rachlinski(2012)]{Wistrich2012z7051}
Andrew~J. Wistrich and Jeffrey~J. Rachlinski.
\newblock {How Lawyers' Intuitions Prolong Litigation Modeling Human
  Decisionmaking in the Law Symposium}.
\newblock \emph{Southern California Law Review}, 86\penalty0 (3):\penalty0
  571--636, 2012.

\bibitem[Zafar et~al.(2017)Zafar, Valera, Gomez~Rodriguez, and
  Gummadi]{Zafar2017z7040}
Muhammad~Bilal Zafar, Isabel Valera, Manuel Gomez~Rodriguez, and Krishna~P.
  Gummadi.
\newblock {Fairness Beyond Disparate Treatment \& Disparate Impact: Learning
  Classification without Disparate Mistreatment}.
\newblock In \emph{Proceedings of the 26th International Conference on World
  Wide Web}, pages 1171--1180, Perth Australia, 2017. International World Wide
  Web Conferences Steering Committee.
\newblock \doi{10.1145/3038912.3052660}.

\end{thebibliography}

\clearpage
\appendix
\section{Empirical Literature}
Empirical literature (including systematic reviews and meta-analyses)
supporting synthetic review.

\begin{longtable}[]{@{}
  >{\raggedright\arraybackslash}p{(\columnwidth - 2\tabcolsep) * \real{0.25}}
  >{\raggedright\arraybackslash}p{(\columnwidth - 2\tabcolsep) * \real{0.75}}@{}}
\toprule\noalign{}
\begin{minipage}[b]{\linewidth}\raggedright
\textbf{Section}
\end{minipage} & \begin{minipage}[b]{\linewidth}\raggedright
\textbf{Studies}
\end{minipage} \\
\midrule\noalign{}
\endhead
\bottomrule\noalign{}
\endlastfoot
Predictive Validity & \citealp{Desmarais2021z7236,Duwe2017z7262,Fazel2012z7230,Fazel2022z7093,Ghasemi2021z7260,Jung2017z7281,Rudin2020z7031} \\
Human-versus-Machine & \emph{Empirical research on human judges}: Abrams
et al., 2012; Ash et al., 2025; Beebe, 2006, 2007, 2020; Bonica \& Sen,
2021; Bontrager et al., 2013; Braman \& Nelson, 2007; Bushway et al.,
2012; Bystranowski et al., 2021; Casper et al., 1989; Chen et al., 2016;
Choi et al., 2022; Cohen \& Yang, 2019; Dhami, 2003; Dhami \& Ayton,
2001; Dyevre \& Rodilla Lazaro, 2026; Eisenberg \& Miller, 2004; Englich
et al., 2006; Epstein et al., 2013; Fessel \& Roese, 2011; Guthrie et
al., 2000; Hangartner et al., 2025; Harley, 2007; Kukucka \& Kassin,
2014; Levinson et al., 2017; Lu \& Chen, 2025; O'Brien, 2009; Philippe,
2020; Pinello, 1999; Rachlinski et al., 2012, 2015; Rachlinski \&
Wistrich, 2021; Ramji-Nogales et al., 2007; Starr, 2015; Sunstein et
al., 2007; Wistrich \& Rachlinski, 2012.

\emph{Human-versus-Machine comparisons}: \citealp{BenMichael2025z7163,Biswas2020z7269,Dressel2018z7032,Kleinberg2018z5972,Lin2020z7165,Rambachan2024z7275,Stevenson2024z7208} \\
Human-plus-Machine & \citealp{Angelova2025z7253,BenMichael2025z7163,Berk2017z7258,Green2019z7194,GrgicHlaca2019z7155,Imai2023z6947,Lawson2024z7263,Viljoen2019z7294} \\
\end{longtable}

\end{document}